\documentclass{article}
\usepackage{iclr2025_conference,times}

\usepackage{amsmath,amsfonts,bm}

\def\eqref#1{equation~\ref{#1}}

\def\1{\bm{1}}

\DeclareMathAlphabet{\mathsfit}{\encodingdefault}{\sfdefault}{m}{sl}
\SetMathAlphabet{\mathsfit}{bold}{\encodingdefault}{\sfdefault}{bx}{n}

\usepackage{hyperref}
\usepackage{url}

\usepackage{amsfonts}
\usepackage{subcaption}
\usepackage{bm}
\usepackage{algorithm}
\usepackage{algorithmic}
\usepackage{amsmath}
\usepackage{enumitem}
\usepackage{makecell}
\usepackage{multirow}
\usepackage{graphicx}
\usepackage{booktabs}

\title{Highly Efficient Self-Adaptive Reward Shaping for Reinforcement Learning}

\author{Haozhe Ma\thanks{The authors contributed equally to this work.} \\
School of Computing \\
National University of Singapore \\
\texttt{haozhe.ma@u.nus.edu} \\
\And
Zhengding Luo\footnotemark[1] \\
School of Electrical and Electronic Engineering~~~~~~~~\\
Nanyang Technological University \\
\texttt{luoz0021@e.ntu.edu.sg} \\
\And
Thanh Vinh Vo \\
School of Computing\\
National University of Singapore \\
\texttt{votv@nus.edu.sg} \\
\And
Kuankuan Sima\\
Department of Electrical and Computer Engineering\\
National University of Singapore \\
\texttt{kuankuan\_sima@u.nus.edu} \\
\And
Tze-Yun Leong \\
School of Computing\\
National University of Singapore \\
\texttt{leongty@nus.edu.sg}
}

\iclrfinalcopy
\begin{document}

\maketitle

\begin{abstract}
    
    Reward shaping is a reinforcement learning technique that addresses the sparse-reward problem by providing frequent, informative feedback. We propose an efficient self-adaptive reward-shaping mechanism that uses success rates derived from historical experiences as shaped rewards. The success rates are sampled from Beta distributions, which evolve from uncertainty to reliability as data accumulates. Initially, shaped rewards are stochastic to encourage exploration, gradually becoming more certain to promote exploitation and maintain a natural balance between exploration and exploitation. We apply Kernel Density Estimation (KDE) with Random Fourier Features (RFF) to derive Beta distributions, providing a computationally efficient solution for continuous and high-dimensional state spaces. Our method, validated on tasks with extremely sparse rewards, improves sample efficiency and convergence stability over relevant baselines.

\end{abstract}

\section{Introduction}
\label{sec:introduction}

Environments with extremely sparse rewards present notable challenges for reinforcement learning (RL). In such contexts, as the reward model lacks immediate signals, agents receive feedback only after long horizons, making the ability to quickly discover beneficial samples crucial for successful learning~\citep{expl-survey:ladosz2022exploration}. To address this, a straightforward solution is to reconstruct the reward models by introducing auxiliary signals that assess the agent's behavior, which has led to the popular technique of Reward Shaping (RS)~\citep{strehl2008analysis,gupta2022unpacking}. Inverse reinforcement learning, which extracts reward functions from human knowledge or expert demonstrations, represents an intuitive approach within this framework~\cite{rs-irl:arora2021survey}. However, it heavily relies on extensive human input, which can be difficult to obtain, especially in complex environments. Alternatively, fully autonomous approaches have emerged as an attractive direction.

Automatically maintained reward shaping can be broadly categorized into two branches: intrinsic motivation-based rewards, which are task-agnostic, and inherent value-based rewards, which are typically task-specific. The former mainly introduces exploration bonuses to encourage agents to explore a wider range of states, commonly by rewarding novel or infrequently visited states~\citep{rs-curi:burda2018exploration,rs-explo:ostrovski2017count,rs-explo:tang2017exploration,rs-explo:bellemare2016unifying}. While these methods effectively enhance exploration, they tend to overlook the internal values of the states. This can lead to the ``noisy TV" problem, where agents fixate on highly novel but meaningless regions, thus trapping them in suboptimal behaviors~\citep{rs-curi:mavor2022stay}. In contrast, the latter leverages high-level heuristics to guide agents in extracting meaningful values from learning experiences, which helps stabilize convergence. However, these methods often struggle in early exploration due to non-directional guidance~\citep{relara:ma2024reward,rs-intr:memarian2021self,trott2019keeping}. 

To overcome the limitations of existing RS methods and combine the advantages of exploration-encouraged and inherent value-based rewards, this paper introduces a novel \textbf{S}elf-\textbf{A}daptive \textbf{S}uccess \textbf{R}ate-based reward shaping mechanism (\textbf{SASR})\footnote{The source code is accessible at: \url{https://github.com/mahaozhe/SASR}}. The success rate, defined as the ratio of a state's presence in successful trajectories to its total occurrences, serves as an auxiliary reward distilled from historical experience. This success rate assesses a state's contribution toward successful task completion, which closely aligns with the agent's original objectives, offering informative guidance for learning. Furthermore, to mitigate overconfidence caused by deterministic success rates, we adopt Beta distributions to model success rates from a probabilistic perspective. Beta distributions enable a self-adaptive evolution of confidence in approximating a state's success rate, ensuring the system gradually converges to reliable rewards as more data is collected, while avoiding premature certainty. To derive Beta distributions, we use kernel density estimation (KDE) with random Fourier features (RFF) to efficiently estimate success and failure counts. The main contributions of this paper are summarized as follows:
\vspace{-\parskip}
\begin{itemize}[noitemsep,leftmargin=*]
\item We propose SASR, an autonomous reward-shaping mechanism for sparse-reward environments. By deriving success rates from historical experiences aligned with the agent's optimization objectives, SASR effectively augments the environmental rewards.
\item We introduce a novel self-adaptive mechanism. Initially, low-confidence Beta distributions provide uncertain rewards, encouraging exploration by perturbing the reward function and assigning higher rewards to unvisited states. As more experience accumulates, high-confidence Beta distributions deliver more reliable and precise rewards to enhance exploitation.
\item To derive Beta distributions in continuous state spaces, we use KDE with RFF, creating an efficient approach that eliminates the need for additional neural networks or models for learning the auxiliary shaped rewards, thereby achieving remarkably low computational complexity.
\item SASR is evaluated on various extremely sparse-reward tasks, significantly outperforming several baselines in sample efficiency, learning speed, and convergence stability.
\end{itemize}

\section{Related Work}

Reward shaping (RS) methods can generally be categorized based on the source of learning: either from \textit{human knowledge} or the \textit{agent's own experiences}. Techniques that derive reward models from human knowledge, such as Inverse Reinforcement Learning (IRL)~\citep{rs-irl:arora2021survey,rs-irl:ramachandran2007bayesian,rs-irl:ziebart2008maximum,rs-irl:hadfield2016cooperative} and Inverse Optimal Control (IOC)~\citep{ioc:schultheis2021inverse,zhang2024scrutinize}, aim to extract reward or objective functions from expert demonstrations. Subsequently, transferring the learned reward models to new tasks has received considerable efforts~\citep{rs-demon:biyik2022learning,rs-imit:wu2021shaping,rs-demon:ellis2021risk,rs-demon:cheng2021heuristic,adamczyk2023utilizing,luo2024gfanc,lyu2024seabo}. However, these methods rely heavily on human-generated data and often struggle adapting out-of-distribution scenarios. Thus, our focus shifts toward autonomous self-learning approaches, which can be further divided into \textit{intrinsic motivation-based} and \textit{inherent value-based} rewards depending on the nature of the rewards.

\textbf{Intrinsic motivation}-based RS explores general heuristics or task-agnostic metrics to encourage exploration. Potential-based algorithms define the shaped reward as $\gamma \Phi(s') - \Phi(s)$, where $\Phi(\cdot)$ is a potential function. This ensures the reward cancels out in the Bellman equation, preserving the optimal policy~\citep{rs-poten:devlin2012dynamic,rs-poten:asmuth2008potential,rs-poten:wiewiora2003potential}. However, designing the potential function highly depends on the environmental dynamics, making it more applicable to model-based RL. More commonly, methods incorporate exploration bonuses to reward novel states~\citep{mahankali2024random,liu2025skill,rs-explo:devidze2022exploration,badia2020never,hong2018diversity,eysenbach2018diversity}. Count-based strategies, for instance, track visitation counts and assign higher rewards to less frequently visited states~\citep{pseudo-count:lobel2023flipping,machado2020count,fox2018dora,fu2017ex2,pseudo-count:martin2017count}. In continuous spaces, state counting is challenging, so \cite{rs-explo:tang2017exploration} introduced a hash function to discretize the state space, \cite{rs-explo:bellemare2016unifying} proposed pseudo-counts based on recording probabilities, and \cite{rs-explo:ostrovski2017count} used PixelCNN~\cite{pixelcnn:van2016conditional} to simulate density. Additionally, random network distillation-based methods measure state novelty by neural networks~\citep{drnd:yang2024exploration,A2PR,rs-curi:burda2018exploration}, while curiosity-driven approaches reward agents for encountering surprising or unpredictable states~\citep{yang2024cmbe,sun2022exploit,burda2019large,rs-curi:pathak2017curiosity,zhang2023how}. Although intrinsic motivation has proven effective in enhancing exploration, only considering novelty while ignoring the inherent values of states can lead to suboptimal policies. 

\textbf{Inherent value} based RS, on the other hand, focuses on task-related signals that highlight how states contribute to achieving higher rewards and their underlying significance. For instance, \cite{trott2019keeping} introduced additional rewards based on the distance between a state and the target; \cite{rs-hier:stadie2020learning} derived informative reward structures using a Self-Tuning Network to optimize guidance; \cite{rs-intr:memarian2021self} captured the preferences among different trajectories by ranking them via a trained classifier; \cite{rs-intr:zheng2018learning} minimized the KL-divergence between learned and original rewards to align their distributions. \cite{rs-multi:mguni2023learning} used an auxiliary agent competing against the original agent in a Markov game; \cite{relara:ma2024reward} introduced ReLara, a collaborative framework where a reward agent automatically generates rewards to guide the policy agent. Moreover, incorporating multiple agents or hierarchical structures to share and transfer knowledge through synchronized reward functions is another promising research direction~\citep{park2023controllability,mine:ma2024mixed,gupta2024behavior,rs-hier:hu2020learning,raileanu2020ride,rs-hier:yi2022learning}. 

\section{Preliminaries}

\noindent\textbf{Reinforcement Learning (RL)} aims to train an agent to interact with an environment, which is commonly modeled as a \textbf{Markov Decision Process (MDP)}. An MDP represented as $\langle S, A, T, R, \gamma \rangle$, involves four main components: $S$ is the state space, $A$ is the action space, $T:S \times A \times S \rightarrow [0,1]$ is the probability of transitioning from one state to another given a specific action, and $R:S \rightarrow \mathbb{R}$ is the reward model. The objective in RL is to learn a policy $\pi(a|s)$ that maximizes the expected cumulative rewards $G = \mathbb{E}[\sum_{t=0}^{\infty} \gamma^t R(s_t)]$, where $\pi(a|s)$ indicates the likelihood of selecting action $a$ in state $s$, and $\gamma$ is the discount factor~\citep{rl:sutton2018reinforcement}.

\noindent\textbf{Beta Distribution} is defined on the interval $[0, 1]$, making it ideal for modeling proportions or probabilities. It is parameterized by $\alpha$ and $\beta$, which represent prior counts of successes and failures of a binary outcome. The probability density function of a Beta-distributed variable $X$ is:
\begin{equation}
\setlength{\abovedisplayskip}{5pt} \setlength{\belowdisplayskip}{5pt}
f(x;\alpha,\beta) = \frac{1}{B(\alpha,\beta)}x^{\alpha-1}(1-x)^{\beta-1},
\end{equation}
where $B(\alpha, \beta)$ is the beta function. The key attribute of Beta distribution is its adaptability: as more data accumulates, the values of $\alpha$ and $\beta$ increase, narrowing the distribution's shape and increasing confidence in the estimated probabilities. This feature is particularly useful in adaptive online learning, aligning with our objective of balancing exploration and exploitation. 

\noindent\textbf{Kernel Density Estimation (KDE)} is a non-parametric method for approximating the probability density function of a random variable from data samples. Given $n$ data points $\{x_i\}_{i=1}^n$, KDE smooths these points to approximate the density function as follows: 
\begin{equation}
\setlength{\abovedisplayskip}{5pt} \setlength{\belowdisplayskip}{5pt}
\hat{d}(x) = \frac{1}{nh}\sum_{i=1}^{n} K\left(\frac{x-x_i}{h}\right),
\end{equation}
where $h$ is the bandwidth, and $K(\cdot)$ is a kernel function such as Gaussian kernel, Laplacian kernel, or Cauchy kernel. KDE is particularly useful in scenarios where the actual distribution is complex or poorly defined, such as in continuous state spaces in RL environments. 

\section{Methodology}

\begin{figure}[t]
\centering
\includegraphics[width=0.98\linewidth]{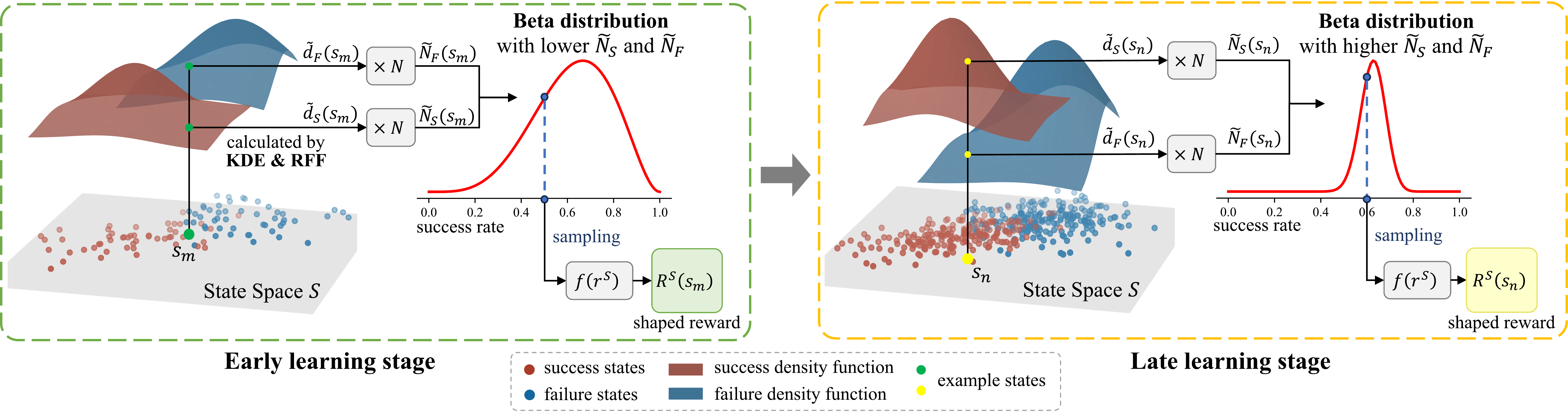}
\caption{A schematic diagram of the self-adaptive success rate based reward shaping mechanism. KDE: Kernel Density Estimation; RFF: Random Fourier Features.}
\label{fig:mechanism}
\vspace{-12pt}
\end{figure}

We propose a \textbf{S}elf-\textbf{A}daptive \textbf{S}uccess \textbf{R}ate based reward shaping mechanism (SASR) to accelerate RL algorithms in extremely-sparse-reward environments. Figure~\ref{fig:mechanism} illustrates the principles of the SASR mechanism: The diagram consists of two parts representing the early and late learning stages. As experiences accumulate with learning progresses, the Beta distributions modeling the success rates evolve from being stochastic to deterministic. This autonomous adaption closely aligns with the agent's exploration-exploitation balance. Section~\ref{sec:meth-sasr} introduces how Beta distributions evolve and how shaped rewards are generated from them. Additionally, to achieve highly efficient computation, we use KDE and RFF to estimate success and failure counts, which are used to derive the corresponding Beta distributions, as detailed in Section~\ref{sec:meth-beta}. Lastly, Section~\ref{sec:meth-algo} presents the integration of SASR into the RL agent and the overall algorithmic flow. 

\subsection{Self-Adaptive Success Rate Sampling}
\label{sec:meth-sasr}

We formulate the augmented reward function in SASR as adding an auxiliary shaped reward $R^S(s)$ to the environmental reward $R^E(s)$, weighting by a factor $\lambda$:
\begin{equation}
R^{SASR}(s) = R^E(s) + \lambda R^S(s).
\end{equation}
We assign the shaped reward $R^S(s_i)$ of a given state based on its \textit{success rate} -- defined as the ratio of the state's presence in successful trajectories to its total occurrences. This metric provides a meaningful reward from a statistical perspective: a higher success rate, reflected in a higher shaped reward, indicates a greater likelihood that the state will guide the agent toward successful task completion. Formally, the \textbf{success rate based shaped reward} $R^S(s_i)$ is given by:
\begin{equation}
R^S(s_i) = f\left(\frac{N_S(s_i)}{N_S(s_i) + N_F(s_i)}\right),
\end{equation}
where $N_S(s_i)$ and $N_F(s_i)$ denote the counts of state $s_i$ appearing in successful and failed historical trajectories, respectively. To enhance scalability and adaptability, $f(\cdot)$ is a linear scaling function that maps the original success rate from $[0, 1]$ to a desired scale $[R_{min}, R_{max}]$, making the magnitude of the shaped rewards more flexible, i.e., $f(x) = R_{min} + x \cdot (R_{max} - R_{min})$.

Given $N_S(s_i)$ and $N_F(s_i)$, directly using a deterministic success rate may lead to overconfidence in the estimation of the true value. To address this, inspired by the principles of Thompson sampling~\citep{thompson-sampling:thompson1933likelihood,thompson-sampling:agrawal2012analysis}, we adopt a probabilistic perspective for success rate estimation. Specifically, the success rate of each state is approximated as a variable in a Beta distribution, with shape parameters set as $\alpha = N_S(s_i) + 1$ and $\beta = N_F(s_i) + 1$:
\begin{align}
r^S_i \sim \text{Beta}(r; \alpha, \beta)
= \frac{1}{B(N_S(s_i) + 1, N_F(s_i) + 1)} r^{N_S(s_i)} (1-r)^{N_F(s_i)},
\end{align}
where the \textit{beta function} $B(\cdot, \cdot)$ is the normalization factor. By sampling from this distribution, we obtain a probabilistic estimate of the true success rate. This sampled value, $r^S_i$, is then processed through the scaling function $f(\cdot)$ to produce the shaped reward: $R^S(s_i) = f(r^S_i)$.

As $N_S(s_i)$ and $N_F(s_i)$ progressively increase throughout the learning process, they influence the shape and sampling variability of the Beta distribution. Generating the shaped reward from these evolving Beta distributions offers several advantages:
\vspace{-\parskip}
\begin{itemize}
[noitemsep,leftmargin=*]
\item \textbf{Encourage Exploration.} In the early phases, lower counts of $N_S(s_i)$ and $N_F(s_i)$ result in higher-variance Beta distributions, making the sampled rewards more stochastic. This acts as a noisy perturbation of the reward function. Given that the environmental rewards are mostly zero, this perturbation optimizes the agent in diverse directions through small adjustments, shifting the anchors from which the stochastic policy samples actions. Meanwhile, early-visited states are likely to fail, leading to a decrease in their success rates, while unvisited states retain the initial Beta distribution $Beta(1,1)$, receiving relatively higher rewards. This mechanism drives the agent to explore novel regions, aligning with the principles of intrinsic motivation.
\item \textbf{Enhance Exploitation.} In the later phases, as the counts $N_S(s_i)$ and $N_F(s_i)$ increase, the Beta distribution gradually sharpens, concentrating generated rewards around the true success rate. The more certain reward signals with higher confidence highly support the agent's exploitation, facilitating faster convergence towards optimal policies.
\item \textbf{Consistent Optimization.} The peak of the Beta distribution, given by $N_S(s_i)/(N_S(s_i) + N_F(s_i))$, equals the success rate. Meanwhile, the expectation, $(N_S(s_i) + 1)/(N_S(s_i) + N_F(s_i) + 2)$, closely approximates the success rate. This ensures that, despite stochasticity, the overall reward remains consistent with policy optimization. 
\end{itemize}

\subsection{Highly Efficient Beta Distribution Derivation}
\label{sec:meth-beta}

In this section, we present how the success and failure counts, $N_S(s_i)$ and $N_F(s_i)$, are derived for the Beta distributions. To efficiently estimate these counts in high-dimensional, continuous, or infinite state spaces, we use Kernel Density Estimation (KDE) to approximate the densities of successes and failures from accumulated experience. Specifically, we maintain two buffers, $\mathcal{D}_S$ and $\mathcal{D}_F$, to store the states in successful and failed trajectories, respectively. By treating these states as scattered data instances distributed across the state space, KDE estimates the density as:
\begin{equation}
\tilde{d}_X(s_i) = \frac{1}{|\mathcal{D}_X|} \sum_{j=1}^{|\mathcal{D}_X|}{K(s_i - s_j)}, \quad X \in \{S, F\},
\end{equation}
where $K(\cdot)$ is the kernel function and $|\mathcal{D}_X|$ is the buffer size. We select Gaussian kernel in our implementation. The estimated density $\tilde{d}_X(s_i)$ approximates the likelihood of encountering state $s_i$ in success or failure scenarios, providing a statistically sound basis for estimating $N_X(s_i)$. By multiplying $\tilde{d}_X(s_i)$ by the total number of observed states $N$, the count $\tilde{N}_X(s_i)$ is estimated as:
\begin{equation}
\tilde{N}_X(s_i) = N \times \tilde{d}_X(s_i) = \frac{N}{|\mathcal{D}_X|} \sum_{j=1}^{|\mathcal{D}_X|}{\exp\left(-\frac{||s_i - s_j||^2}{2 h^2}\right)}, \quad X \in \{S, F\},
\end{equation}
where hyperparameter $h$ is the bandwidth of the Gaussian kernel.

We further integrate Random Fourier Features (RFF)~\citep{rff:rahimi2007random} to reduce computational complexity, as calculating the Gaussian kernel can be expensive, especially in scenarios involving high-dimensional state spaces and large buffers. RFF approximates the kernel function of the original $k$-dimensional states through an inner product of $M$-dimensional randomized features:
\begin{equation}
K(s_i, s_j) \approx z(s_i)^T z(s_j), \quad z(s) = \sqrt{\frac{2}{M}} \cos(\bm{W}^T s + \bm{b}),
\end{equation}
where $z(\cdot)$ is the RFF mapping function with $\bm{W} = \big[\bm{w}^{(1)}, \ldots, \bm{w}^{(M)} \big] \in \mathbb{R}^{k \times M}$, and $\bm{b} = \big[b^{(1)}, \ldots, b^{(M)} \big]^T \in \mathbb{R}^M$ is randomly sampled from the following  distributions:
\begin{equation}
\label{eq:rff-sampling}
\bm{w}^{(m)} \sim \mathcal{N}(\bm{0}, \sigma^{-2}\bm{I}_k), \quad b^{(m)} \sim \text{Uniform}(0, 2\pi), \quad m = 1, \ldots, M, 
\end{equation}
where $\bm{I}_k$ is the $k \times k$ identity matrix. Equation~\ref{eq:rff-sampling} is applied for the Gaussian kernel, while different kernels and the detailed derivations of the RFF method are provided in Appendix~\ref{sec:appendix-rff-derivation}.

\subsubsection{Implementation Details}

\textbf{Retention Rate.} We introduce a hyperparameter, the \textit{retention rate} $\phi \in (0,1]$, to regulate the volume and diversity of states stored in the buffers. Rather than storing all encountered states, we uniformly retain a specific portion of $\phi$. The motivations behind this are: (1) adjacent states in one trajectory tend to be highly similar, especially those near the initial state are repetitive and uninformative, retaining a portion of states can skip redundant states and increase sample diversity; (2) using a lower retention rate in the early stage keeps $N_S$ and $N_F$ lower, resulting in broader Beta distributions and preventing premature overconfidence.

\textbf{Defining Success and Failure.} In tasks where sparse rewards are only given at the end of an episode to indicate task completion, the entire trajectory can be classified as either a success or failure based on the episodic reward. For tasks with sparse rewards that do not explicitly indicate task completion, we segment the trajectories by positive reward occurrences. Specifically, if a reward is obtained within a pre-defined maximum steps, the corresponding sub-sequence is classified as a success; otherwise, it is considered a failure.

\subsubsection{Time and Space Complexity of SASR}

Suppose the buffer size of $\mathcal{D}_X$ is $D$ and the batch size is $B$ per iteration, the computational complexity to compute the count $N_X$ is $O(MDB)$, indicating linear complexity (detailed in Appendix~\ref{sec:appendix-complexity}). RFF converts nonlinear kernel computations into linear vector operations, significantly speeding up computation by leveraging the vectorization capabilities of GPUs~\citep{dongarra2014accelerating}. This highlights its superior efficiency compared to methods that rely on network updates and inferences involving extensive nonlinear computations. 

Given the retention rate $\phi$, the space complexity of maintaining two buffers is $O(\phi T |s|)$, where $T$ is the total iterations and $|s|$ is the size of a single state. Moreover, storage space is significantly reduced by leveraging the existing replay buffer in off-policy RL algorithms, like SAC~\citep{sac:haarnoja2018soft} and TD3~\citep{td3:fujimoto2018addressing}. Specifically, we augment the replay buffer with a flag that marks each state as either a success ($\text{flag}=1$) or failure ($\text{flag}=0$). This allows for the efficient management of space requirements with minimal overhead from indexing.

For supporting experimental results in time and space complexity, please refer to Appendix~\ref{sec:time-space-complexity}.

\subsection{The SASR Mechanism for RL agents}
\label{sec:meth-algo}

\begin{algorithm}[t]
\caption{Self-Adaptive Success Rate based Reward Shaping}
\label{alg:sasr}
\begin{algorithmic}[1]
\REQUIRE Environment $\mathcal{E}$ and agent $\mathcal{A}$.
\REQUIRE Experience replay buffer $\mathcal{D}$.
\REQUIRE State buffers for success $\mathcal{D}_S$ and failure $\mathcal{D}_F$.
\REQUIRE RFF mapping function $z: \mathbb{R}^k \rightarrow \mathbb{R}^M$. \hfill $\triangleright$ Sample $\bm{W}$ and $\bm{b}$ based on Equation~\ref{eq:rff-sampling}

\FOR{each trajectory $\tau = \emptyset$}
    \FOR{each environmental step}
        \STATE $(s_t, a_t, s_{t+1}, r^E_t) \leftarrow \text{CollectTransition}(\mathcal{E}, \mathcal{A})$ \hfill $\triangleright$ interact with the environment
        \STATE $\mathcal{D} \leftarrow \mathcal{D} \cup \{(s_t, a_t, s_{t+1}, r^E_t)\}$ \hfill $\triangleright$ store the transition in the replay buffer
        \STATE $\tau \leftarrow \tau \cup \{s_t\}$ \hfill $\triangleright$ record the state in the trajectory
    \ENDFOR

    \STATE \textbf{if} trajectory is successful: $\mathcal{D}_S \leftarrow \mathcal{D}_S \cup \tau$ \hfill $\triangleright$ store the trajectory in the success buffer
    \STATE \textbf{else}: $\mathcal{D}_F \leftarrow \mathcal{D}_F \cup \tau$ \hfill $\triangleright$ otherwise, store the trajectory in the failure buffer
\ENDFOR

\FOR{each update step}
    \STATE $\{(s_t, a_t, r^E_t, s_{t+1})_i\} \sim \mathcal{D}$ \hfill $\triangleright$ sample a batch of transitions from the replay buffer
    \STATE $\tilde{N}_S = z(s_t)^T z(\mathcal{D}_S)$ \hfill $\triangleright$ estimate the success counts
    \STATE $\tilde{N}_F = z(s_t)^T z(\mathcal{D}_F)$ \hfill $\triangleright$ estimate the failure counts
    \STATE $r^S_t \sim \text{Beta}(r; \tilde{N}_S + 1, \tilde{N}_F + 1)$ \hfill $\triangleright$ sample the success rate from the Beta distribution
    \STATE $r^{SASR}_t = r^E_t + \lambda f(r^S_t)$ \hfill $\triangleright$ compute the SASR reward
    \STATE Update agent $\mathcal{A}$ with $\{(s_t, a_t, r^{SASR}_t, s_{t+1})_i\}$
\ENDFOR
\end{algorithmic}
\end{algorithm}

Building upon the SASR reward, for demonstration, we employ the soft actor-critic (SAC) algorithm by~\citet{sac:haarnoja2018soft} as the foundational agent. Let $Q_{\psi}$ be the parameterized Q-network and $\pi_{\theta}$ be the parameterized policy network. The Q-network is optimized by the following loss function:
\begin{equation}
\label{eq:loss-qf}
\mathcal{L}(\psi) = \Big(Q_{\psi}(s_t, a_t) - \big( r^E_t + \lambda R^S(s_t) + \gamma Q_{{\psi}'}(s_{t+1}, a_{t+1}) \big) \Big)^2,
\end{equation}
where $Q_{{\psi}'}$ is obtained from a secondary frozen target network to maintain a fixed objective~\citep{dqn:mnih2015human}. Notably, the environmental reward $r^E_t$ is retrieved from the replay buffer, conversely, the shaped reward $R^S(s_t)$ is computed in real-time using the most recently updated $N_S(s_t)$ and $N_F(s_t)$, ensuring it reflects the latest learning progress.

We optimize the policy network by maximizing the expected Q-value and the policy entropy $\mathcal{H}\big(\pi_{\theta}(\cdot | s_t)\big)$, following~\cite{sac-ap:haarnoja2018soft}:
\begin{equation}
\label{eq:loss-pi}
\mathcal{L}(\theta) = \mathbb{E}_{a_t \sim \pi_{\theta}(\cdot | s_t)}\big[-Q_{\psi}(s_t, a_t) + \log{\pi_{\theta}(a_t | s_t)}\big].
\end{equation}

The flow of the SAC-embedded SASR algorithm is summarized in Algorithm~\ref{alg:sasr}. 

\section{Experiments}

We evaluate SASR in high-dimensional environments, including four \textit{MuJoCo} tasks~\citep{env:todorov2012mujoco}, four robotic tasks~\citep{env:gymnasium_robotics2023github}, five \textit{Atari} games, including the well-known \textit{Montezuma's Revenge}~\cite{env:bellemare2013arcade}, and a physical simulation task~\citep{env:towers_gymnasium_2023}, as shown in Figure~\ref{fig:tasks}. All tasks provide extremely sparse rewards, with a reward of $1$ granted only upon reaching the final objective within the maximum permitted steps. To ensure robust validation, we run $10$ instances per setting with different random seeds and report the average results. We also maintain consistent hyperparameters and network architectures across all tasks, detailed in Appendix~\ref{sec:appendix-network-parameters}.

\begin{figure}[h]
\centering
\includegraphics[width=0.72\linewidth]{Images/environments.pdf}
\vspace{-8pt}
\caption{\textit{MuJoCo}, robotic, \textit{Atari} games and physical simulation tasks in our experiments. Detailed descriptions and the environmental reward models of each task are provided in Appendix~\ref{sec:appendix-environments}.}
\label{fig:tasks}
\vspace{-6pt}
\end{figure}

\subsection{Comparison and Discussion}
\label{sec:comparison}

\noindent\textbf{Baselines.} We compare SASR with ten baselines to benchmark its performance: (a) the online Distributional Random Network Distillation (DRND)~\citep{drnd:yang2024exploration}, (b) RL with an Assistant Reward Agent (ReLara)~\citep{relara:ma2024reward}, (c) General Function Approximation Reward-Free Exploration (GFA-RFE)~\citep{gfa:zhang2024uncertainty}, (d) RL Optimizing Shaping Algorithm (ROSA)~\citep{rs-multi:mguni2023learning}, (e) Exploration-Guided RS (ExploRS)~\citep{rs-explo:devidze2022exploration}, (f) Count-based static hashing exploration (\#Explo)~\citep{rs-explo:tang2017exploration}, (g) Random Network Distillation (RND)~\citep{rs-curi:burda2018exploration}, (h) Soft Actor-Critic (SAC)~\citep{sac:haarnoja2018soft}, (i) Twin Delayed DDPG (TD3)~\citep{td3:fujimoto2018addressing}, and (j) Proximal Policy Optimization (PPO)~\citep{ppo:schulman2017proximal}. Algorithms (a) to (g) are all reward shaping methods, incorporating either exploration bonuses or auxiliary agents to shape rewards, while algorithms (h) to (j) are advanced RL algorithms.

\begin{figure}
\centering
\includegraphics[width=0.92\linewidth]{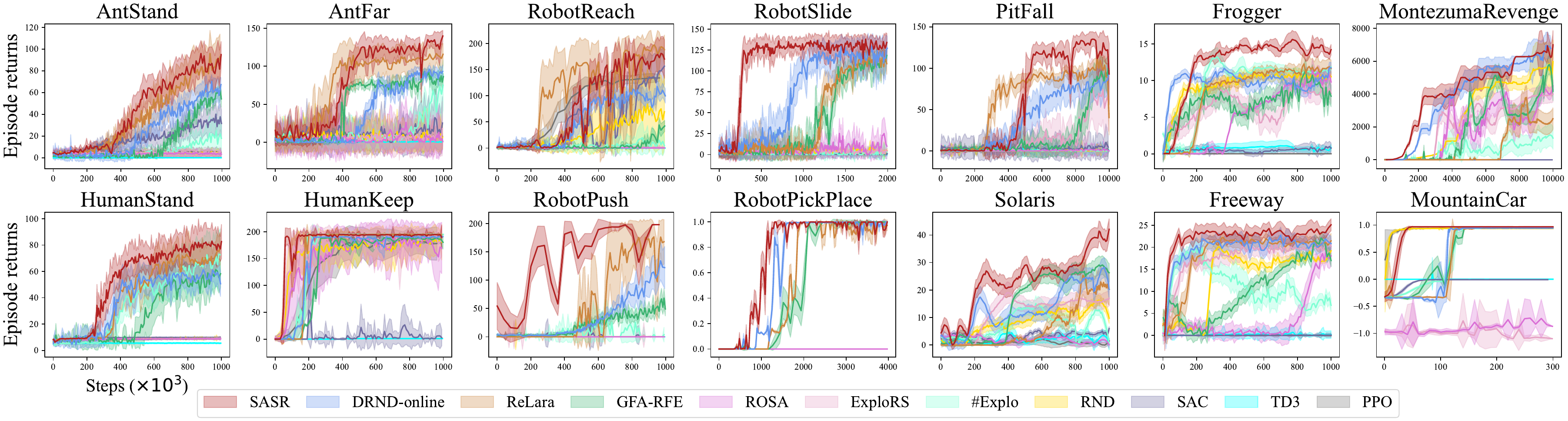}
\vspace{-8pt}
\caption{The learning performance of SASR compared with the baselines.}
\label{fig:comparison-baselines}
\vspace{-8pt}
\end{figure}

Figure~\ref{fig:comparison-baselines} shows the learning performance of SASR compared with the baselines, while Table~\ref{tab:comparison-baselines} reports the average episodic returns with standard errors achieved by the final models over $100$ episodes. Our findings indicate that SASR surpasses the baselines in terms of sample efficiency, learning stability, and convergence speed. The primary challenge in these environments is the extremely sparse reward given after a long horizon, making exploration crucial for obtaining successful trajectories in a timely manner. Although exploration strategies of algorithms such as ExploRS, \#Explo, and RND are designed to reward novel states, effectively expanding the early exploration space with direct additional targets, they continue prioritizing novelty, overlooking the implicit values of these states. As a result, they fail to return to the final objectives. 

\begin{table}
\centering
\tiny
\setlength{\tabcolsep}{1.7pt}
\caption{The average episodic returns and standard errors of all models tested over 100 episodes.}
\vspace{-6pt}
\label{tab:comparison-baselines}

\begin{tabular}{cccccccccccc}
\toprule
Tasks & SASR & DRND-online & ReLara & GFA-RFE & ROSA & ExploRS & \#Explo & RND & SAC & TD3 & PPO \\
\midrule
AntStand & \textbf{94.9$\pm$0.0} & 67.3$\pm$0.0 & 90.5$\pm$1.7 & 54.2$\pm$0.0 & 3.8$\pm$0.4 & 5.1$\pm$0.4 & 17.9$\pm$0.0 & 4.0$\pm$0.2 & 31.6$\pm$0.0 & 0.0$\pm$0.0 & 4.9$\pm$0.1 \\
AntFar & \textbf{139.8$\pm$0.0} & 93.2$\pm$0.0 & 115.7$\pm$0.0 & 86.4$\pm$0.0 & 1.0$\pm$0.0 & 12.0$\pm$4.2 & 75.1$\pm$0.0 & 4.6$\pm$1.6 & 25.3$\pm$0.0 & 1.0$\pm$0.0 & 7.8$\pm$0.0 \\
HumanStand & \textbf{79.8$\pm$2.0} & 50.6$\pm$0.0 & 76.2$\pm$0.7 & 58.2$\pm$0.0 & 8.8$\pm$0.0 & 9.3$\pm$0.0 & 72.7$\pm$0.0 & 9.3$\pm$0.1 & 9.9$\pm$0.0 & 5.5$\pm$0.0 & 9.0$\pm$0.1 \\
HumanKeep & \textbf{195.8$\pm$0.0} & 154.5$\pm$0.0 & 194.9$\pm$0.0 & 141.5$\pm$0.0 & 169.7$\pm$0.0 & 182.8$\pm$0.0 & 195.0$\pm$0.0 & 180.7$\pm$0.0 & 2.5$\pm$0.0 & 1.0$\pm$0.0 & 138.1$\pm$0.0 \\
RobotReach & 170.2$\pm$0.0 & 99.8$\pm$0.0 & \textbf{187.9$\pm$0.0} & 42.1$\pm$0.0 & 0.1$\pm$0.0 & 0.7$\pm$0.0 & 4.6$\pm$0.0 & 69.3$\pm$0.0 & 156.5$\pm$0.0 & 0.0$\pm$0.0 & 79.5$\pm$0.0 \\
RobotSlide & \textbf{132.3$\pm$1.3} & 127.2$\pm$0.0 & 111.6$\pm$2.0 & 115.8$\pm$2.0 & 11.2$\pm$0.9 & 4.3$\pm$0.1 & 3.5$\pm$0.0 & 4.8$\pm$0.2 & 0.7$\pm$0.2 & 0.5$\pm$0.4 & 0.2$\pm$0.2 \\
RobotPush & \textbf{167.1$\pm$0.0} & 122.2$\pm$0.0 & 166.9$\pm$0.0 & 49.1$\pm$0.0 & 0.0$\pm$0.0 & 0.0$\pm$0.0 & 3.7$\pm$0.0 & 0.0$\pm$0.0 & 0.0$\pm$0.0 & 0.0$\pm$0.0 & 0.0$\pm$0.0 \\
RobotPickPlace & \textbf{1.0$\pm$0.0} & \textbf{1.0$\pm$0.0} & 1.0$\pm$0.0 & 0.5$\pm$0.0 & 0.0$\pm$0.0 & 0.0$\pm$0.0 & 0.0$\pm$0.0 & 0.0$\pm$0.0 & 0.0$\pm$0.0 & 0.0$\pm$0.0 & 0.0$\pm$0.0 \\
Pitfall & \textbf{93.0$\pm$0.0} & 92.0$\pm$0.0 & 40.3$\pm$0.0 & 89.4$\pm$0.0 & 0.0$\pm$0.0 & 57.6$\pm$0.0 & 0.0$\pm$0.0 & 0.0$\pm$0.0 & 4.6$\pm$0.0 & 0.5$\pm$0.0 & 0.0$\pm$0.0 \\
Frogger & \textbf{14.2$\pm$0.0} & 11.7$\pm$0.0 & 11.6$\pm$0.0 & 7.9$\pm$0.0 & 9.8$\pm$0.0 & 8.3$\pm$0.0 & 11.9$\pm$0.0 & 10.5$\pm$0.0 & 0.8$\pm$0.0 & 0.7$\pm$0.0 & 0.0$\pm$0.0 \\
Montezuma & 6737.9$\pm$0.0 & \textbf{6828.5$\pm$0.0} & 2421.9$\pm$0.0 & 4755.3$\pm$0.0 & 4294.4$\pm$0.0 & 3971.5$\pm$0.0 & 1400.1$\pm$0.0 & 5494.3$\pm$0.0 & 0.0$\pm$0.0 & 0.0$\pm$0.0 & 0.0$\pm$0.0 \\
Solaris & \textbf{42.1$\pm$0.0} & 21.3$\pm$0.7 & 20.3$\pm$0.0 & 26.3$\pm$0.0 & 0.1$\pm$0.0 & 17.0$\pm$0.0 & 1.2$\pm$0.8 & 9.8$\pm$0.0 & 6.0$\pm$0.0 & 0.4$\pm$0.0 & 1.5$\pm$0.0 \\
Freeway & \textbf{22.4$\pm$0.0} & 19.8$\pm$0.0 & 21.5$\pm$0.0 & 10.1$\pm$0.0 & 18.0$\pm$0.0 & 17.5$\pm$0.0 & 6.9$\pm$0.0 & 13.0$\pm$0.0 & 0.1$\pm$0.0 & 0.2$\pm$0.0 & 0.0$\pm$0.0 \\
MountainCar & \textbf{1.0$\pm$0.0} & \textbf{1.0$\pm$0.0} & \textbf{1.0$\pm$0.0} & \textbf{1.0$\pm$0.0} & -0.9$\pm$0.0 & -1.0$\pm$0.0 & \textbf{1.0$\pm$0.0} & \textbf{1.0$\pm$0.0} & -0.1$\pm$0.0 & 0.0$\pm$0.0 & 0.9$\pm$0.0 \\
\bottomrule
\end{tabular}
\vspace{-10pt}
\end{table}

SASR outperforms the baselines primarily due to its self-adaptive reward evolution mechanism. In the early phases, SASR encourages exploration by injecting substantial random rewards, optimizing the agent in multiple directions, and increasing the likelihood of collecting positive samples. Moreover, since most states are initially classified as failures, their success rates decrease. As a result, unvisited states receive relatively higher rewards, encouraging further exploration. This mechanism resembles intrinsic motivation, assigning higher rewards to novel states, and effectively guiding the agent to expand the exploration space.
As more data is collected, the success rate estimation becomes more accurate, the shaped reward provides more reliable guidance, enhancing exploitation and stabilizing convergence. Together, these strategies improve SASR's sample efficiency and convergence stability in challenging tasks.


While ReLara used a similar exploration mechanism by perturbing reward functions, it relies on an independent black-box agent, requiring more iterations to converge. In contrast, SASR's success rate sampling is more direct, reducing delays in acquiring valuable information. ReLara's advantage lies in incorporating the policy agent's actions into reward construction, as seen in the \textit{RobotReach} task, where the target point is randomly selected in each episode. In this case, ReLara outperforms SASR due to access to action information. However, SASR can achieve the same by incorporating actions as additional features in the state vector.

\subsection{Effect of Self-Adaptive Success Rate Sampling}
\label{sec:exp-balance}

\begin{figure*}[h]
\centering
\includegraphics[width=0.88\linewidth]{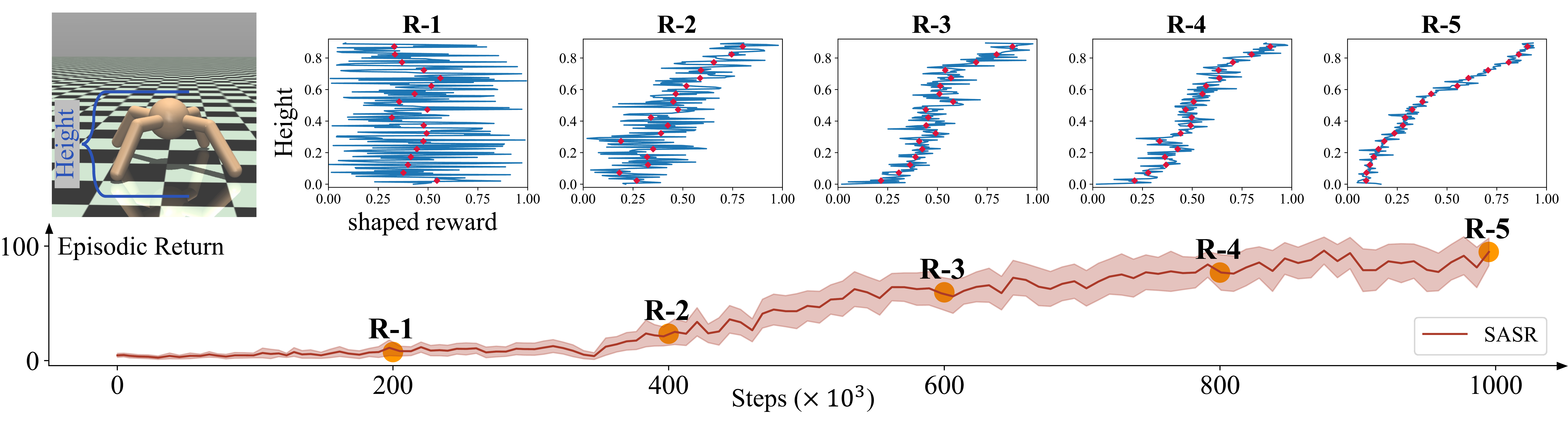}
\captionsetup{aboveskip=2pt}
\caption{Distributions of the shaped rewards over the height of the ant robot in the \textit{AntStand} task at different training stages. Red diamonds represent the estimated success rate, while the blue polylines show the actual shaped rewards sampled from the Beta distribution.}
\label{fig:self-ee-balance}
\vspace{-10pt}
\end{figure*}

SASR introduces a novel self-adaptive mechanism that balances exploration and exploitation by modulating the randomness of the shaped rewards. To further investigate the effect of this mechanism, we use the \textit{AntStand} task as a case study, analyzing the shaped rewards learned at different training stages. Figure~\ref{fig:self-ee-balance}~(bottom) shows the learning curve, while Figure~\ref{fig:self-ee-balance}~(top) illustrates the distributions of generated rewards over the ``height of the ant" feature, a dimension in the state vector. 

\begin{figure*}
\centering
\includegraphics[width=0.82\linewidth]{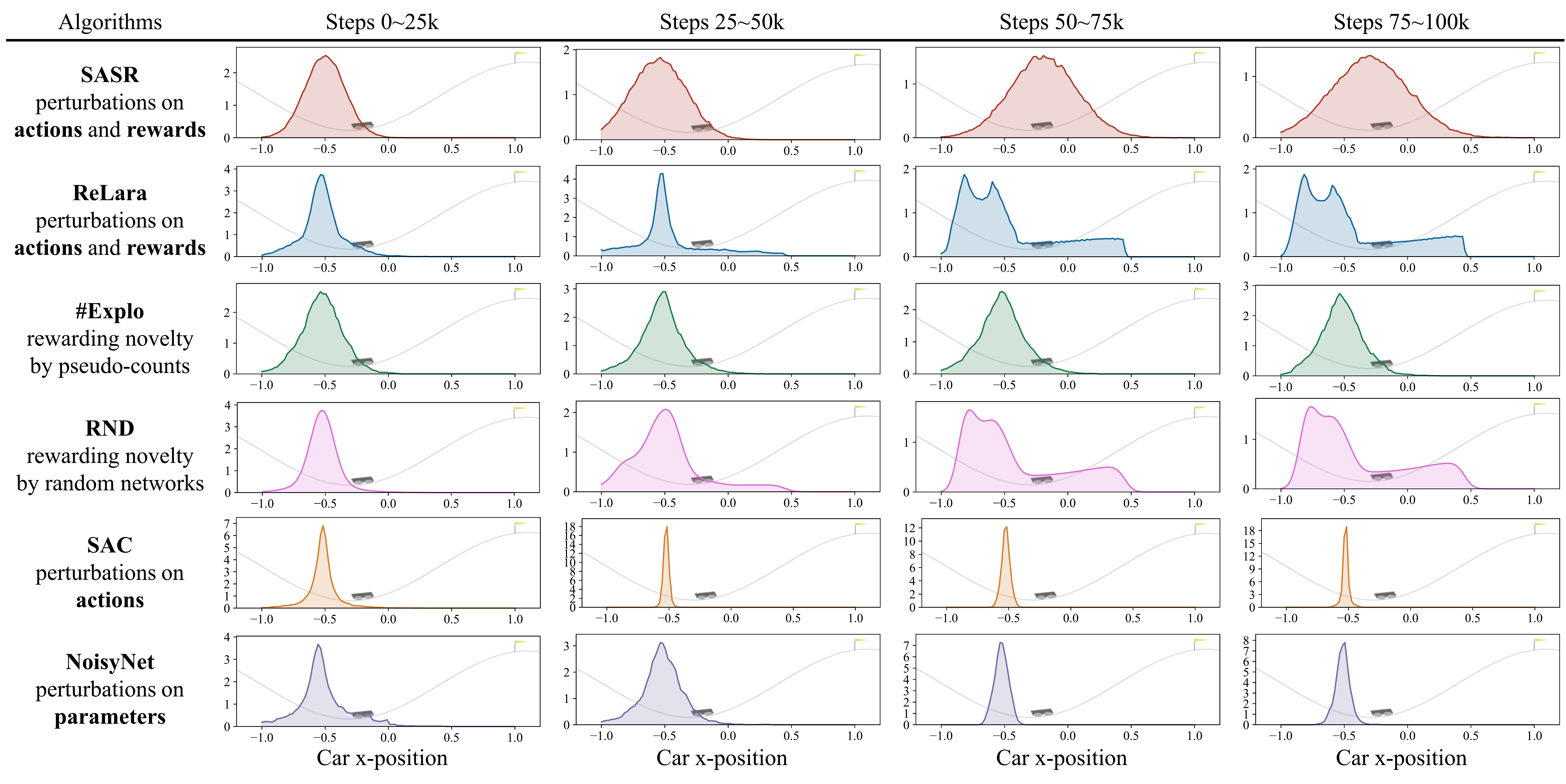}
\captionsetup{aboveskip=4pt}
\caption{The density of visited states in the \textit{MountainCar} task for four training periods.}
\label{fig:visit-density}
\vspace{-16pt}
\end{figure*}

As learning progresses, the shaped rewards exhibit two key attributes: values transition from random to meaningful, and variance decreases from uncertain to deterministic. Although the sampled rewards fluctuate, their means gradually show a positive linear correlation with the robot's height. In the early phases, the shaped rewards contain significant randomness due to high uncertainty. While these random signals offer limited information, they drive the agent to take small optimization steps in diverse directions, effectively shifting the policy anchors, expanding the action space sampled from SAC's stochastic policy, promoting exploration, and increasing sample diversity. In later phases, rewards stabilizes, closely aligning with the robot's height -- a meaningful and intuitive metric -- enhancing the agent's exploitation. 

To further investigate SASR's exploration behavior, we compare the visited state density throughout training in the \textit{MountainCar} task with five representative exploration strategies: (1) ReLara, which perturbs both rewards and actions; (2) \#Explo and (3) RND, which rewards novel states; (4) SAC, which uses entropy-regularized exploration; and (5) NoisyNet~\citep{fortunato2018noisy}, which perturbs network weights. The state density for every 25k steps is shown in Figure~\ref{fig:visit-density}. We observe that SASR progressively covers a wider range of the state space. From 50k to 100k steps, SASR reaches positions near the goal, driven by its success rate mechanism. In contrast, ReLara and RND cover similar ranges to SASR, but are less smooth and take longer to reach the right side. \#Explo shows no clear rightward shift, as it prioritizes novelty and ignores the inherent value of states. SAC's exploration is relatively narrow, making it prone to getting trapped in local optima. NoisyNet's range narrows over time as perturbations diminish through optimization. Overall, SASR demonstrates more effective exploration and collects valuable samples sooner, leading to faster convergence.

\subsection{Ablation Study}

We conduct ablation studies to investigate key components of SASR. We select six representative tasks and report the experimental results, with quantitative data provided in Appendix~\ref{sec:appendix-ablation-study}. 

\begin{figure*}[t]
    \centering
    \begin{subfigure}[t]{0.9\linewidth}
        \centering
        \includegraphics[width=\linewidth]{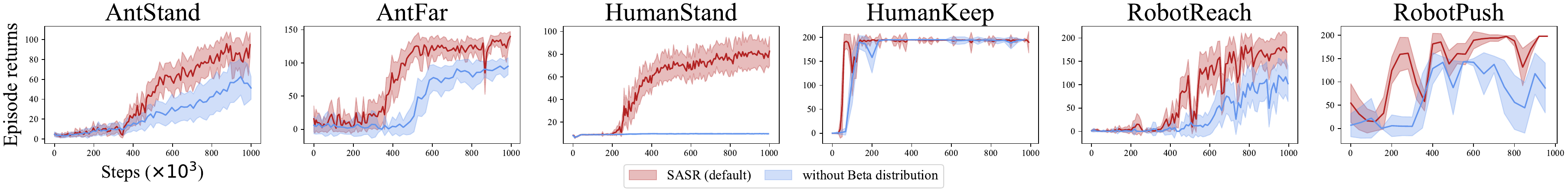}
        \captionsetup{aboveskip=1pt, belowskip=3pt}
        \caption{The impact of omitting Beta distribution sampling on the performance of SASR.}
        \label{fig:abla-woBeta}
    \end{subfigure}
    
    \begin{subfigure}[t]{0.9\linewidth}
        \centering
        \includegraphics[width=\linewidth]{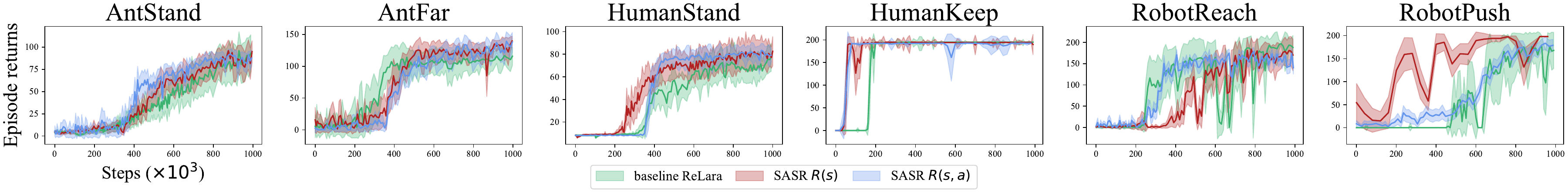}
        \captionsetup{aboveskip=1pt, belowskip=3pt}
        \caption{Learning performance of SASR with shaped reward function $R^S(s,a)$ and $R^S(s)$.}
        \label{fig:abla-rsa-vs-rs}
    \end{subfigure}

    \begin{subfigure}[t]{0.9\linewidth}
        \centering
        \includegraphics[width=\linewidth]{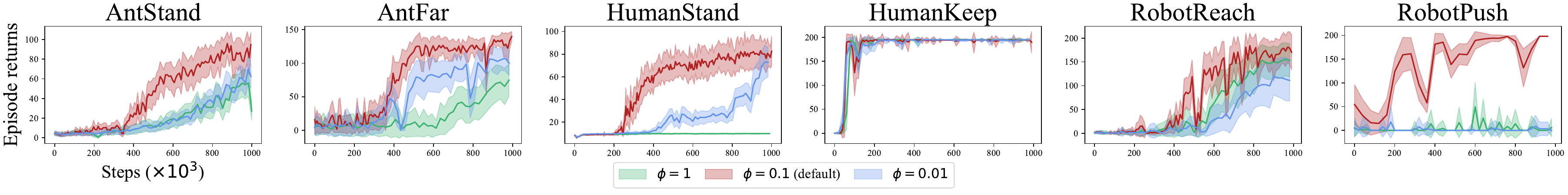}
        \captionsetup{aboveskip=1pt, belowskip=3pt}
        \caption{Learning performance of SASR with different retention rates $\phi$.}
        \label{fig:abla-diff-rr}
    \end{subfigure}
    
    \begin{subfigure}[t]{0.9\linewidth}
        \centering
        \includegraphics[width=\linewidth]{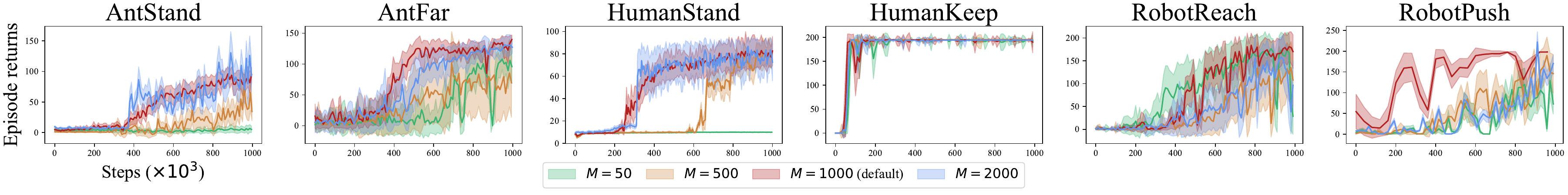}
        \captionsetup{aboveskip=1pt, belowskip=3pt}
        \caption{Learning performance of SASR with different RFF feature dimensions $M$.}
        \label{fig:abla-rff-dim}
    \end{subfigure}
    
    \begin{subfigure}[t]{0.9\linewidth}
        \centering
        \includegraphics[width=\linewidth]{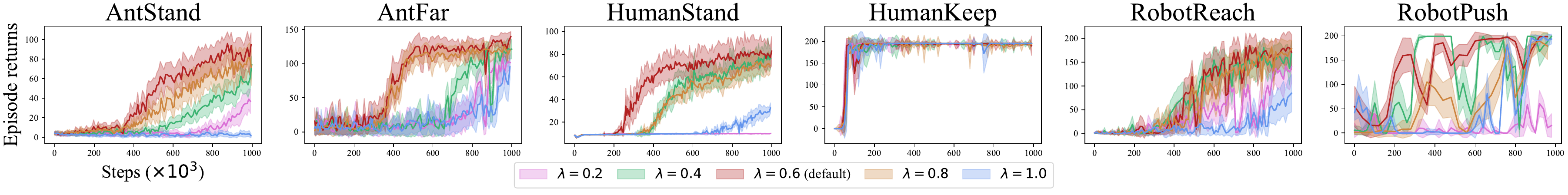}
        \captionsetup{aboveskip=1pt, belowskip=3pt}
        \caption{Learning performance of SASR with different shaped reward weight factors $\lambda$.}
        \label{fig:abla-diff-beta}
    \end{subfigure}
    
    \captionsetup{aboveskip=2pt}
    \caption{Ablation study: the impact of key components on the performance of SASR.}
    \label{fig:overall-ablation}
\vspace{-6pt}
\end{figure*}

\noindent \textbf{Sampling from Beta distributions.} (Figure~\ref{fig:abla-woBeta}) We examine a variant of SASR that omits Beta distribution sampling, instead directly using the success rate $N_S(s_i)/(N_S(s_i) + N_F(s_i))$. In the early stages, limited experience makes this success rate an unreliable estimate, and using a fixed, overly confident value can mislead the agent. Furthermore, skipping Beta distribution sampling eliminates exploration driven by random rewards, leading to narrower exploration. The results highlight the critical role of Beta distribution sampling in effective learning.

\noindent \textbf{Reward function over state-action pair.} (Figure~\ref{fig:abla-rsa-vs-rs}) We extend SASR with a reward function over state-action pairs, $r(s, a)$. The comparison results show that both settings perform similarly. However, encoding actions into the reward function increases dimensionality, complicating density estimation and correlation assessment. Furthermore, the state and action vectors may have different distributions, potentially reducing KDE estimation accuracy.

\noindent \textbf{Retention rate $\phi$.} (Figure~\ref{fig:abla-diff-rr}) The retention rate directly influences the confidence level of Beta distributions. A high retention rate ($\phi=1$) preserves all samples, resulting in a densely populated, redundant state pool, which makes the Beta distribution prematurely overconfident and degrades performance. Conversely, a low retention rate ($\phi=0.01$) slows convergence as more iterations are required to gather sufficient samples. The results suggest that an appropriate retention rate is crucial.

\noindent \textbf{RFF feature dimensions $M$.} (Figure~\ref{fig:abla-rff-dim}) SASR shows relatively low sensitivity to $M$, provided it is sufficiently large to capture the complexity of the states. Results show that values like $M=500, 1000, 2000$ all yield similar performance, while significantly lower dimensions, such as $M=50$, degrade performance. These findings align with the original RFF study~\citep{rff:rahimi2007random}.

\noindent \textbf{Shaped reward weight factor $\lambda$.} (Figure~\ref{fig:abla-diff-beta}) SASR performs better with intermediate values like $\lambda=0.4, 0.6, 0.8$. At $\lambda=0.2$, the minimal shaped reward scale reduces state differentiation, leading to suboptimal performance. At $\lambda=1$, aligning the shaped reward scale with the environmental reward introduces excessive uncertainty and randomness, potentially overwhelming feedback and hindering learning. The findings emphasize that maintaining a balanced reward scale is important for optimal learning outcomes.

\section{Conclusion and Discussion}
\label{sec:conclusion}

In this paper, we propose SASR, a self-adaptive reward shaping algorithm based on success rates to address the sparse-reward challenge. SASR achieves a balance between exploration and exploitation by generating shaped rewards from evolving Beta distributions. Experiments demonstrate that this adaptability significantly improves the agent's convergence speed. Additionally, the use of KDE and RFF provides a computationally efficient approach to deriving Beta distributions. This mechanism also offers a sound alternative to traditional count-based RS strategies, adapting effectively to continuous environments. Our evaluations confirm the superior performance of SASR in terms of sample efficiency and learning stability.

While SASR is designed for sparse-reward environments, in dense-reward settings, the additional shaped rewards may be unnecessary. Extending SASR to such scenarios presents a promising direction for further research. Moreover, the derivation of Beta distributions relies on the samples stored in the success and failure buffers. Currently, our method does not consider the relationships or varying importance of different states within the same trajectory, making it sensitive to the retention rate. Therefore, developing an adaptive retention rate or improved buffer management mechanisms is crucial for future improvement.

\newpage
\section*{Acknowledgement}

This research is supported by the National Research Foundation Singapore and DSO National Laboratories under the AI Singapore Programme (Award Number: AISG2-RP-2020-016). This research is also partially supported by an Academic Research Grant No. MOE-T2EP20121-0015 from the Ministry of Education in Singapore.

\bibliography{reference}
\bibliographystyle{iclr2025_conference}

\newpage
\appendix
\section{Appendix}

\subsection{Derivation of Random Fourier Features}
\label{sec:appendix-rff-derivation}

We incorporate Random Fourier Features (RFF)~\citep{rff:rahimi2007random} to approximate the kernel functions in the KDE process for the SASR algorithm. Let the original state be $k$-dimensional, denoted as $\bm{s} \in \mathbb{R}^k$, and the kernel function be $k(\bm{s}_i, \bm{s}_j)$. RFF approximates the kernel function by projecting the input $k$-dimensional space into a $M$-dimensional feature space using a mapping function $z: \mathbb{R}^k \rightarrow \mathbb{R}^M$. The RFF-based kernel function is then defined as follows:
\begin{equation}
k(\bm{s}_i, \bm{s}_j) \approx z(\bm{s}_i)^T z(\bm{s}_j),
\end{equation}
We provide the derivation of the RFF approximation in this section.

First, we clarify that RFF primarily targets shift-invariant kernels, that satisfy $k(\bm{s}_i, \bm{s}_j) = k(\bm{s}_i-\bm{s}_j)$. Common shift-invariant kernels include Gaussian kernels, Laplacian kernels, and Cauchy kernels. Given a shift-invariant kernel function $k(\Delta)$, we perform the inverse Fourier transform:
\begin{align}
k(\bm{s}_i, \bm{s}_j) &= \int_{\mathbb{R}^k} p(\bm{w}) e^{\mathsf{i} \bm{w}^T (\bm{s}_i - \bm{s}_j)} d\bm{w} \\
&= \mathbb{E}_{\bm{w}}\big[e^{\mathsf{i} \bm{w}^T (\bm{s}_i - \bm{s}_j)}\big],
\end{align}
where we can consider $\bm{w} \sim p(\bm{w})$ based on the Bochner's theorem, and $p(\bm{w})$ is called the \textit{spectral distribution} of kernel function. For the three types of shift-invariant kernels, the corresponding spectral distributions are listed in Table~\ref{tab:appendix-spectral}:

\begin{table}[h]
\caption{Some shift-invariant kernels and their associated spectral distributions.}
\label{tab:appendix-spectral}
\centering
\setlength{\tabcolsep}{5pt}
\begin{tabular}{@{}lcc@{}}
\toprule
\textbf{Kernel} & \makecell[c]{\textbf{Kernel function}, $\displaystyle k(\bm{s}_i - \bm{s}_j)$} & \makecell[c]{\textbf{Spectral density}, $\displaystyle p(\bm{w})$} \\ 
\midrule
Gaussian & $\displaystyle\exp\left(-\frac{\|\bm{s}_i - \bm{s}_j\|_2^2}{h^2}\right)$ & $\displaystyle\frac{\sqrt{h}}{2\sqrt{\pi}}\exp\left(-\frac{h\|\bm{w}\|_2^2}{4}\right)$ \\
Laplacian & $\displaystyle\exp\big(-\|\bm{s}_i-\bm{s}_j\|_1\big)$ & $\displaystyle \prod_{m=1}^M\frac{1}{\pi(1 + w_d^2)}$ \\ 
Cauchy & $\displaystyle \prod_{i=1}^k\frac{2}{\pi\big(1 + (\bm{s}_i-\bm{s}_j)^2\big)}$ & $\displaystyle\exp\big(-\|\bm{w}\|_1\big)$ \\ 
\bottomrule
\end{tabular}
\end{table}

Next, we perform the Euler's formula transformation, which retains only the cosine term since we are dealing with real-valued functions, the kernel function can be further derived as:
\begin{align}
k(\bm{s}_i, \bm{s}_j) &= \mathbb{E}_{\bm{w}}\big[e^{\mathsf{i} \bm{w}^T (\bm{s}_i - \bm{s}_j)}\big] \\
&= \mathbb{E}_{\bm{w}}\big[\cos(\bm{w}^T (\bm{s}_i - \bm{s}_j))\big] \\
&= \mathbb{E}_{\bm{w}}\big[\cos(\bm{w}^T (\bm{s}_i - \bm{s}_j))\big] + \mathbb{E}_{\bm{w}}\big[\mathbb{E}_{b}[\cos(\bm{w}^T (\bm{s}_i + \bm{s}_j) + 2b)]\big] \label{eq:derive-1} \\ 
&= \mathbb{E}_{\bm{w}}\big[\mathbb{E}_{b}[\cos(\bm{w}^T (\bm{s}_i - \bm{s}_j)) + \cos(\bm{w}^T (\bm{s}_i + \bm{s}_j) + 2b)]\big] \\ 
&= \mathbb{E}_{\bm{w}}\big[\mathbb{E}_{b}[\sqrt{2}\cos(\bm{w}^T \bm{s}_i + b) \sqrt{2}\cos(\bm{w}^T \bm{s}_j + b)]\big], \label{eq:derive-2}
\end{align}
where $b \sim \text{Uniform}(0, 2\pi)$. Equation~\ref{eq:derive-1} holds since $\mathbb{E}_{b \sim \text{Uniform}(0, 2\pi)}\big[\cos(t + 2b)\big] = 0$ for any $t$. Equation~\ref{eq:derive-2} is  obtained from $\cos(A-B) + \cos(A+B) = 2 \cos(A)\cos(B)$, where $A=\bm{w}^T\bm{s}_i + b$, $B=\bm{w}^T\bm{s}_j + b$.

We define the mapping $z_{\bm{w}, b}(\bm{s}) = \sqrt{2}\cos(\bm{w}^T \bm{s} + b)$, then the kernel function can be approximated by the inner product of two vectors and the expectation can be approximated by Monte Carlo sampling:
\begin{align}
k(\bm{s}_i, \bm{s}_j) &= \mathbb{E}_{\bm{w}}\big[\mathbb{E}_{b}[z_{\bm{w}, b}(\bm{s}_i) z_{\bm{w}, b}(\bm{s}_j)]\big] \\
&\approx \frac{1}{M} \sum_{m=1}^{M} z_{\bm{w}_d, b_d}(\bm{s}_i) z_{\bm{w}_d, b_d}(\bm{s}_j) \\
&= z(\bm{s}_i)^T z(\bm{s}_j).
\end{align}

Therefore, we have derived the mapping function $z(\bm{s}) = \sqrt{2/M} \cos(\bm{W}^T \bm{s} + \bm{b})$, where $\bm{W} \in \mathbb{R}^{M \times k}$ and $\bm{b} \in \mathbb{R}^M$. The RFF-based kernel function can be approximated by the inner product of the mapped features in the $M$-dimensional space.

\subsection{Computational Complexity}
\label{sec:appendix-complexity}

In this section, we derive the computational complexity to retrieve the success or failure counts $N_S$ and $N_F$ for each iteration. Suppose the buffer size of $\mathcal{D}_X$ is $D$, the batch size of $\mathcal{B}$ is $B$, the corresponding counts are retrieved by calculating:
\begin{equation}
\tilde{N}_X = N \times z(\bm{\mathcal{B}})^T z(\bm{\mathcal{D}}_X),
\end{equation}
where the mapping function is defined as:
\begin{equation}
z(\bm{s}) = \sqrt{\frac{2}{M}} \cos(\bm{W}^T \bm{s} + \bm{b}), \quad \bm{W}\in \mathbb{R}^{k \times M}, \quad \bm{b} \in \mathbb{R}^M.
\end{equation}

For each state, the mapping function calculation involves:
\begin{enumerate}
[noitemsep,leftmargin=*]
\item Matrix multiplication $\bm{W}^T \bm{s}$: $kM$.
\item Addition $\bm{W}^T \bm{s} + \bm{b}$: $M$.
\item Cosine calculation $\cos(\bm{W}^T \bm{s} + \bm{b})$: $M$.
\end{enumerate}
Therefore, the computational complexity for calculating $z(\bm{s})$ for one state is $O(kM)$.

For each pair of states $(\bm{s}_i,\bm{s}_j)$, calculating the kernel involves $M$ multiplications and $M-1$ additions, thus, the complexity is $O(M)$.

For each iteration, we calculate the RFF mapping for all states in the buffer and the batch and then compute the kernel between them. The complexities involve three parts: RFF mapping for the buffer, RFF mapping for the batch and kernel calculation:
\begin{equation}
O(DkM) + O(BkM) + O(MDB).
\end{equation}
Since the first two terms $O(DkM)$ and $O(BkM)$ are dominated by the last term $O(MDB)$ when the buffer size and the batch size are large, the overall computational complexity to retrieve the corresponding counts can be approximated as $O(MDB)$.

\subsection{Experiments on Time and Space Complexity}
\label{sec:time-space-complexity}

\subsubsection{Time and Space Complexity Comparison}
\label{sec:appendix-complexity-analysis}

In this section, we analyze the time and space overhead introduced by SASR and other representative reward-shaping methods. Below, we summarize the computational and memory costs of the RS baselines, introduced by the shaped reward generation.
\vspace{-\parskip}
\begin{itemize}[noitemsep,leftmargin=*]
    \item \textbf{SASR (ours)} calculates shaped rewards using RFF, which essentially is matrix operations, without additional networks/models learning processes. Regarding the memory costs, the buffers $D_S$ and $D_F$ are much smaller than the replay buffer used in the backbone SAC algorithm, due to the retention rate $\phi$. While considering the scalability for larger problems, we have implemented an alternative approach by augmenting the original replay buffer in the backbone SAC algorithm with a success or failure flag. This approach avoids the need for additional buffers.
    \item \textbf{ReLara}~\citep{relara:ma2024reward} requires an additional RL agent (of the same scale as the original RL agent) and an additional replay buffer.
    \item \textbf{ROSA}~\citep{rs-multi:mguni2023learning} involves a competition agent (the same sacle as the original RL agent) and a switching model (a neural network).
    \item \textbf{ExploRS}~\citep{rs-explo:devidze2022exploration} requires learning two parameterized networks: one for a self-supervised reward model and another for the exploration bonus.
    \item \textbf{\#Explo}~\citep{rs-explo:tang2017exploration} requires a hash function to discretize the state space and a hash table to store the state-visitation counts.
    \item \textbf{RND}~\citep{rs-curi:burda2018exploration} uses a random network distillation module to compute the intrinsic rewards.  
\end{itemize}

Furthermore, we report the computational and memory costs of SASR and the RS baselines in two tasks: \textit{AntStand} and \textit{Frogger}, the results are shown in Table~\ref{tab:appendix-time-comp} and Table~\ref{tab:appendix-space-comp}, respectively. To provide a more intuitive comparison, we report the relative value normalized to our SASR, in this case, if the value $>1$, it indicates that the baseline is more computationally or memory expensive than SASR, and vice versa.

\begin{table}[h]
\centering
\caption{Average maximum memory consumption during the training process, normalized to SASR.}
\label{tab:appendix-time-comp}
\begin{tabular}{cccccccc}
\toprule
Tasks & \underline{SASR} & ReLara & ROSA & ExploRS & \#Explo & RND \\
\midrule
\textit{AntStand} & \underline{1} & 3.67 & 4.12 & 2.05 & 0.89 & 0.12 \\
\textit{Frogger} & \underline{1} & 5.21 & 4.33 & 2.64 & 0.92 & 0.09 \\
\bottomrule
\end{tabular}
\end{table}

\begin{table}[h]
\centering
\caption{Average training time, normalized to SASR.}
\label{tab:appendix-space-comp}
\begin{tabular}{cccccccc}
\toprule
Tasks & \underline{SASR} & ReLara & ROSA & ExploRS & \#Explo & RND \\
\midrule
\textit{AntStand} & \underline{1} & 1.87 & 2.12 & 1.67 & 1.08 & 1.11 \\
\textit{Frogger} & \underline{1} & 1.98 & 3.17 & 1.72 & 1.24 & 1.06 \\
\bottomrule
\end{tabular}
\end{table}

\subsubsection{Comparison of SASR with and without RFF}
\label{sec:appendix-training-time}

To evaluate the effect of introducing RFF, we compare the training time of SASR with and without RFF, also with the backbone SAC algorithm, the results are shown in Table~\ref{tab:appendix-time}. The tests are conducted on the NVIDIA RTX A6000 GPUT. The results show that excluding SAC's inherent optimization time, RFF significantly saves time in the SASR algorithm, while with varying effects across tasks.

\begin{table}[h]
\centering
\caption{Comparison of training time (in hours) for SASR with and without RFF.}
\label{tab:appendix-time}
\setlength{\tabcolsep}{4.5pt}
\begin{tabular}{cccccccc}
\toprule
Algorithms & \textit{AntStand} & \textit{AntFar} & \textit{HumanStand} & \textit{HumanKeep} & \textit{RobotReach} & \textit{RobotPush} \\
\midrule
SAC (backbone) & 5.87 & 5.08 & 4.87 & 5.67 & 5.42 & 6.3 \\
SASR KDE+RFF & 7.15 & 7.52 & 6.92 & 6.20 & 7.07 & 8.13 \\
SASR w/o RFF & 8.12 & 8.72 & 8.37 & 6.53 & 11.12 & 9.21 \\
\bottomrule
\end{tabular}
\end{table}

\subsection{Auto Hyperparameter Selection}

To improve the robustness and generalization of the SASR algorithm, we propose some potential autonomous hyperparameter selection strategies, mainly designed for the bandwidth $h$ of the kernel function and the RFF feature dimension $M$. 

For bandwidth $h$, we can use the empirical formula \textit{Silverman's Rule of Thumb}~\citep{book:wilcox2011introduction}:
\begin{equation}
    h=1.06 \cdot \sigma \cdot N^{-1/5},    
\end{equation}
or cross-validation to determine the optimal bandwidth.

For the RFF dimension $M$, it is directly related to the bandwidth $h$. After determining $h$, we can use the formula mentioned in the RFF theory to determine $M$: 
\begin{equation}
    M = O(\frac{1}{\epsilon^2} \log{\frac{N}{\delta}}),    
\end{equation}
where $\epsilon$ and $\delta$ are the error and confidence parameters. Another method is to compare the Frobenius norm error between the RFF approximated kernel matrix $K^{RFF}$ and the true kernel matrix $K^{Gaussian}$ to select $M$: $\|K^{RFF} - K^{Gaussian}\|_F$.

\subsection{Supplementary Experimental Results for Ablation Study}
\label{sec:appendix-ablation-study}

In this section, we provide the detailed quantitative results of the ablation study.
\begin{table}[!h]
\centering
\small
\caption{Ablation study \#1: The average episodic returns and standard errors of SASR and the variant without sampling from Beta distributions.}
\label{tab:appendix-woBeta}

\begin{tabular}{ccc}
\toprule
Tasks & SASR (with sampling) & SASR (without sampling) \\
\midrule
\textit{AntStand} & \textbf{94.92$\pm$0.00} & 54.48 $\pm$ 1.29  \\
\textit{AntFar} & \textbf{139.84$\pm$0.00} & 92.77 $\pm$ 1.53 \\
\textit{HumanStand} & \textbf{79.83$\pm$2.03} & 9.77 $\pm$ 0.02 \\
\textit{HumanKeep} & \textbf{195.77$\pm$0.00} & 185.00 $\pm$ 0.00 \\
\textit{RobotReach} & \textbf{170.18$\pm$0.00} & 110.29 $\pm$ 2.93 \\
\textit{RobotPush} & \textbf{167.14$\pm$0.00} & 86.82 $\pm$ 0.00 \\
\bottomrule
\end{tabular}
\end{table}

\begin{table}[!h]
\centering
\small
\caption{Ablation study \#2: The average episodic returns and standard errors of SASR with reward function on state-action pair or state only.}
\label{tab:appendix-diff-rwf}
\begin{tabular}{ccc}
\toprule
Tasks & SASR (with $R^S(s)$) & SASR (with $R^S(s,a)$) \\
\midrule
\textit{AntStand} & \textbf{94.92$\pm$0.00} & 85.61$\pm$1.30 \\
\textit{AntFar} & \textbf{139.84$\pm$0.00} & 132.49$\pm$2.92 \\
\textit{HumanStand} & \textbf{79.83$\pm$2.03} & 78.93$\pm$0.65 \\
\textit{HumanKeep} & \textbf{195.77$\pm$0.00} & 192.54$\pm$0.16 \\
\textit{RobotReach} & \textbf{170.18$\pm$0.00} & 151.95$\pm$5.74 \\
\textit{RobotPush} & 167.14$\pm$0.00 & \textbf{179.76$\pm$1.66} \\
\bottomrule
\end{tabular}
\end{table}

\begin{table}[!h]
\centering
\small
\caption{Ablation study \#3: The average episodic returns and standard errors of SASR with different retention rates.}
\label{tab:appendix-diff-rr}
\begin{tabular}{cccc}
\toprule
Tasks & $\phi=1$ & $\phi=0.1$ (default) & $\phi=0.01$ \\
\midrule
\textit{AntStand} & 45.71$\pm$7.57 & \textbf{94.92$\pm$0.00} & 62.85$\pm$3.49 \\
\textit{AntFar} & 70.07$\pm$3.30 & \textbf{139.84$\pm$0.00} & 103.65$\pm$2.57 \\
\textit{HumanStand} & 9.88$\pm$0.01 & \textbf{79.83$\pm$2.03} & 66.46$\pm$2.96 \\
\textit{HumanKeep} & 195.00$\pm$0.00 & \textbf{195.77$\pm$0.00} & 194.77$\pm$0.10 \\
\textit{RobotReach} & 154.32$\pm$0.89 & \textbf{170.18$\pm$0.00} & 112.46$\pm$0.90 \\
\textit{RobotPush} & 2.96$\pm$1.97 & \textbf{167.14$\pm$0.00} & 1.75$\pm$1.24 \\
\bottomrule
\end{tabular}
\end{table}

\begin{table}[!h]
\centering
\small
\caption{Ablation study \#4: The average episodic returns and standard errors of SASR with different RFF feature dimensions $M$.}
\label{tab:appendix-diff-M}

\begin{tabular}{ccccc}
\toprule
Tasks & $M=50$ & $M=500$ & $M=1000$ (default) & $M=2000$ \\
\midrule
\textit{AntStand} & 5.21$\pm$0.45 & 50.68$\pm$6.40 & 94.92$\pm$0.00 & \textbf{96.80$\pm$8.42} \\
\textit{AntFar} & 98.88$\pm$2.64 & 72.17$\pm$5.07 & \textbf{139.84$\pm$0.00} & 129.87$\pm$0.63 \\
\textit{HumanStand} & 9.87$\pm$0.01 & 78.82$\pm$0.52 & \textbf{79.83$\pm$2.03} & 77.73$\pm$1.47 \\
\textit{HumanKeep} & 193.84$\pm$0.61 & 194.71$\pm$0.15 & 195.77$\pm$0.00 & \textbf{195.86$\pm$0.03} \\
\textit{RobotReach} & 119.09$\pm$26.92 & 122.89$\pm$4.51 & \textbf{170.18$\pm$0.00} & 94.87$\pm$15.56 \\
\textit{RobotPush} & 71.67$\pm$27.53 & 161.70$\pm$11.54 & \textbf{167.14$\pm$0.00} & 150.20$\pm$8.70 \\
\bottomrule
\end{tabular}    
\end{table}

\begin{table}[!h]
\centering
\small
\caption{Ablation study \#5: The average episodic returns and standard errors of SASR with different shaped reward weight factors.}
\label{tab:appendix-diff-beta}
\setlength{\tabcolsep}{4.5pt}
\begin{tabular}{cccccc}
\toprule
Tasks & $\lambda=0.2$ & $\lambda=0.4$ & $\lambda=0.6$ (default) & $\lambda=0.8$ & $\lambda=1.0$ \\
\midrule
\textit{AntStand} & 35.71$\pm$0.92 & 59.82$\pm$2.71 & \textbf{94.92$\pm$0.00} & 75.61$\pm$1.41 & 3.16$\pm$0.35 \\
\textit{AntFar} & 99.83$\pm$3.25 & 119.82$\pm$1.18 & \textbf{139.84$\pm$0.00} & 119.35$\pm$1.80 & 80.71$\pm$4.74 \\
\textit{HumanStand} & 9.81$\pm$0.02 & 75.35$\pm$1.06 & \textbf{79.83$\pm$2.03} & 70.96$\pm$0.51 & 28.94$\pm$0.46 \\
\textit{HumanKeep} & 194.68$\pm$0.08 & 194.21$\pm$0.38 & \textbf{195.77$\pm$0.00} & 193.85$\pm$0.42 & 194.89$\pm$0.10 \\
\textit{RobotReach} & 131.61$\pm$4.51 & 154.16$\pm$5.20 & \textbf{170.18$\pm$0.00} & 169.00$\pm$2.56 & 74.23$\pm$4.23 \\
\textit{RobotPush} & 13.07$\pm$1.65 & \textbf{193.56$\pm$3.85} & 167.14$\pm$0.00 & 178.90$\pm$0.00 & 192.07$\pm$0.87 \\
\bottomrule
\end{tabular}    
\end{table}

\textbf{Bandwidth $h$ of Gaussian kernel.} The bandwidth $h$ controls the smoothness of the kernel functions. Beyond fixed bandwidths, we also test a linearly decreasing configuration ($h:0.5 \rightarrow 0.1$), which reflects increasing confidence in KDE. Results indicate that a small bandwidth ($h=0.01$) increases the distance between samples, causing many to have zero estimated density, while a large bandwidth ($h = 1$) makes samples indistinguishable due to an overly flat kernel function. Both cases result in suboptimal performance. The decreasing bandwidth setting offers no significant improvement and tends to reduce stability due to inconsistent density estimations.

\begin{table}[!h]
\centering
\small
\caption{Ablation study \#6: The average episodic returns and standard errors of SASR with different bandwidths $h$ of Gaussian kernel.}
\label{tab:appendix-diff-h}
\setlength{\tabcolsep}{4.5pt}
\begin{tabular}{cccccc}
\toprule
Tasks & $h=0.01$ & $h=0.1$ & $h=0.2$ (default) & $h=1$ & $h=0.5\rightarrow 0.1$ \\
\midrule
\textit{AntStand} & 10.71$\pm$2.52 & 57.22$\pm$3.87 & \textbf{94.92$\pm$0.00} & 17.74$\pm$2.53 & 68.40$\pm$1.54 \\
\textit{AntFar} & 17.58$\pm$2.84 & 99.80$\pm$4.41 & \textbf{139.84$\pm$0.00} & 25.83$\pm$8.62 & 136.49$\pm$4.15 \\
\textit{HumanStand} & 9.89$\pm$0.01 & 64.47$\pm$1.87 & \textbf{79.83$\pm$2.03} & 9.90$\pm$0.02 & 58.79$\pm$2.76 \\
\textit{HumanKeep} & 194.92$\pm$0.02 & 194.00$\pm$0.57 & \textbf{195.77$\pm$0.00} & 193.06$\pm$0.46 & 194.59$\pm$0.18 \\
\textit{RobotReach} & 128.57$\pm$3.83 & 97.35$\pm$19.12 & \textbf{170.18$\pm$0.00} & 134.02$\pm$2.02 & 59.39$\pm$26.02 \\
\textit{RobotPush} & 2.29$\pm$1.62 & 122.45$\pm$37.58 & \textbf{167.14$\pm$0.00} & 0.00$\pm$0.00 & 0.01$\pm$0.01 \\
\bottomrule
\end{tabular}    
\end{table}

\subsection{Network Structures and Hyperparameters}
\label{sec:appendix-network-parameters}

\subsubsection{Network Structures}

\begin{figure}[h!]
    \centering
    \includegraphics[width=0.75\linewidth]{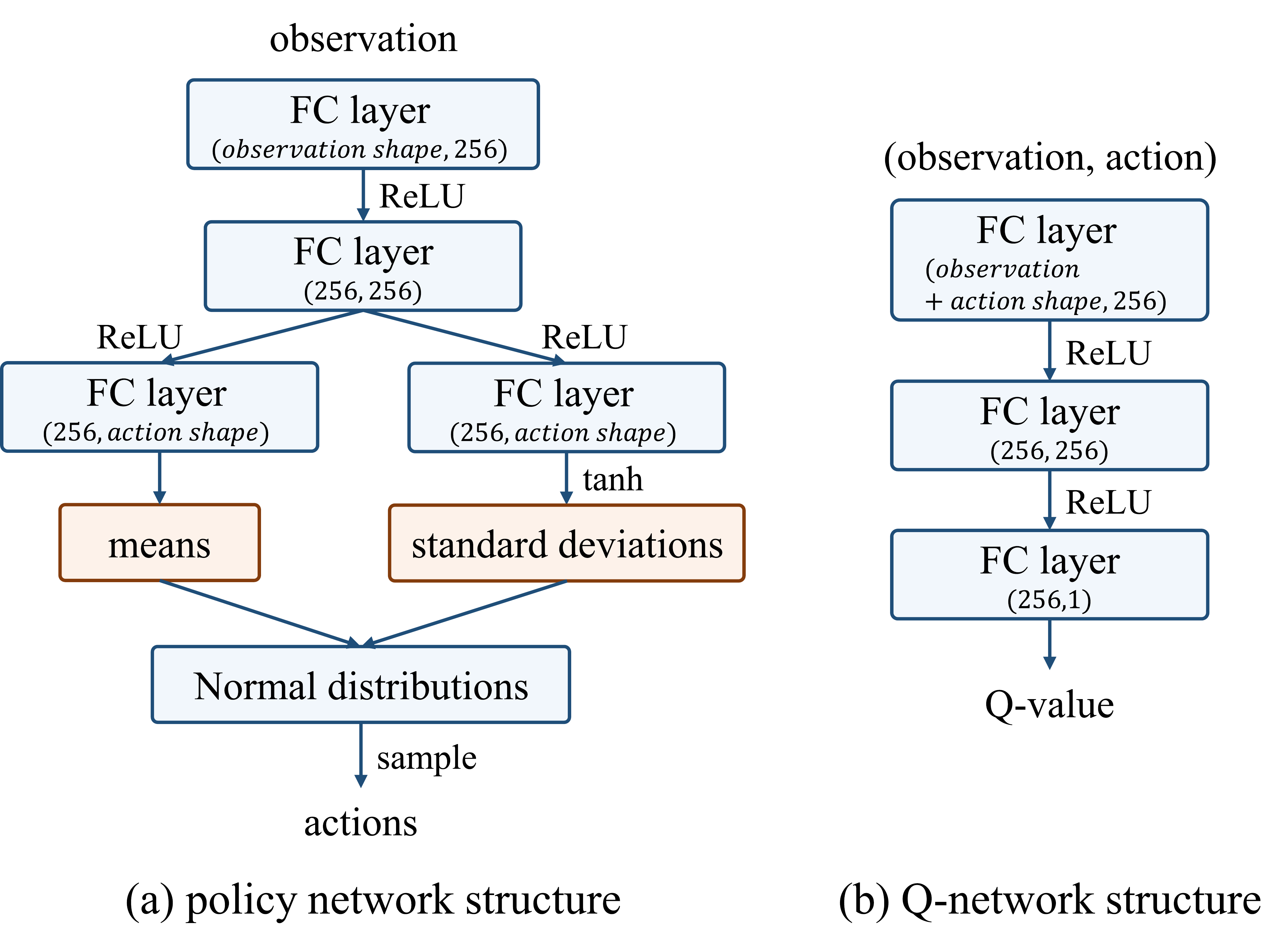}
    \caption{The structures of policy network and Q-network in our implementation.}
    \label{fig:appendix-networks}
    \vspace{-10pt}
\end{figure}

Figure~\ref{fig:appendix-networks} illustrates the structures of the policy network and Q-network employed in our experiments. The agent utilizes simple multilayer perceptron (MLP) models for these networks. Given the use of stochastic policies, the policy network features separate heads to generate the means and standard deviations of the inferred normal distributions, which are then used to sample actions accordingly.

\subsubsection{Hyperparameters}

We have observed that SASR demonstrated high robustness and was not sensitive to hyperparameter choices. Table~\ref{tab:appendix-hyperparameters} shows the set of hyperparameters that we used in all of our experiments.

\begin{table*}[h!]
\centering        
\small
\caption{The hyperparameters used in the SASR algorithm.}
\label{tab:appendix-hyperparameters}

\begin{tabular}{cc}
    \toprule
    Hyperparameters & Values \\
    \midrule
    reward weight $\lambda$ (default) & 0.6 \\
    kernel function bandwidth & 0.2 \\
    random Fourier features dimension $M$ & 1000 \\
    retention rate $\phi$ (default) & 0.1 \\
    \midrule
    discounted factor $\gamma$ & 0.99 \\
    replay buffer size $|\mathcal{D}|$ & $1 \times 10^6$ \\ 
    batch size & 256 \\
    actor module learning rate & $3 \times 10^{-4}$ \\
    critic module learning rate & $1 \times 10^{-3}$ \\
    SAC entropy term factor $\alpha$ learning rate & $1 \times 10^{-4}$ \\
    policy networks update frequency (steps) & $2$ \\
    target networks update frequency (steps) & $1$ \\
    target networks soft update weight $\tau$ & $5 \times 10^{-3}$ \\
    burn-in steps & $5000$ \\
    \bottomrule
\end{tabular}
\end{table*}

\subsubsection{Compute Resources}
\label{sec:appendix-resources}

The experiments in this paper were conducted on a computing cluster, with the detailed hardware configurations listed in Table~\ref{tab:appendix-resources}. The computing time for the SASR algorithm in each task (running 1,000,000 steps) was approximately $6 \pm 2$ hours.

\begin{table*}[h]
\centering
\small
\caption{The compute resources used in the experiments}
\label{tab:appendix-resources}

\begin{tabular}{cc}
    \toprule
    Component & Specification \\
    \midrule
    Operating System (OS) & Ubuntu 20.04 \\
    Central Processing Unit (CPU) & 2x Intel Xeon Gold 6326 \\
    Random Access Memory (RAM) & 256GB \\
    Graphics Processing Unit (GPU) & 1x NVIDIA A100 20GB \\
    Brand & Supermicro 2022 \\
    \bottomrule
\end{tabular}
\end{table*}

\subsection{Configurations of Tasks}
\label{sec:appendix-environments}

In this section, we provide the detailed configurations of the tasks in the experiments.

\begin{itemize}[noitemsep,leftmargin=*]
    \item \textit{AntStand}: The ant robot is trained to stand over a target position. The reward is given if the ant robot reaches the target height. Maximum episode length is 1000 steps.
    \item \textit{AntFar}: The ant robot is trained to reach a target position far from the starting point. The reward is given if the ant robot reaches the target position. Maximum episode length is 1000 steps.
    \item \textit{HumanStand}: The human robot is trained to stand over a target position. The robot is initialized by lying on the ground, and the reward is given if the robot reaches the target height. Maximum episode length is 1000 steps.
    \item \textit{HumanKeep}: The human robot is trained to keep a target height. The robot is initialized by standing, and the reward is given if the robot maintains the target height. Maximum episode length is 1000 steps.
    \item \textit{RobotReach}: The robot arm is trained to reach a target position. The target position is randomly generated in the workspace, and the reward is given if the robot reaches the target position. Maximum episode length is 500 steps.
    \item \textit{RobotPush}: The robot arm is trained to push an object to a target position. The target position is randomly generated on the table, and the reward is given if the object reaches the target position. Maximum episode length is 500 steps.
    \item \textit{RobotSlide}: The robot arm is trained to slide an object to a target position. The target position is randomly generated on the table, and the reward is given if the object reaches the target position. Maximum episode length is 500 steps.
    \item \textit{RobotPickPlace}: The robot arm is trained to pick and place an object to a target position. The target position is randomly generated in the space, and the reward is given if the object reaches the target position. Maximum episode length is 500 steps.
    \item \textit{Pitfall}: The agent is tasked with collecting all the treasures in a jungle while avoiding the pitfalls. The reward is given if the agent collects one treasure, while if the agent falls into a pitfall, the episode ends. Maximum episode length is 2000 steps.
    \item \textit{Frogger}: The agent is trained to cross frogs on a river. The reward is given when each frog is crossed, and the episode ends if all frogs are crossed or fall into the river. Maximum episode length is 2000 steps.
    \item \textit{MontezumaRevenge}: The agent is trained to navigate through a series of rooms to collect keys and reach the final room. The reward is given if the agent successfully reaches one new room. Maximum episode length is 5000 steps.
    \item \textit{Solaris}: The agent controls a spaceship to blast enemies and explore new galaxies. The reward is given if the agent destroys one enemy spaceship and enters a new galaxy. Maximum episode length is 2000 steps.
    \item \textit{Freeway}: The agent is trained to guide the chicken across multiple lanes of heavy traffic. The reward is given if one chicken crosses one lane, while the episode ends if all chickens are crossed or hit by a car. Maximum episode length is 2000 steps.
    \item \textit{MountainCar}: The car is trained to reach the top of the right hill. The reward is given if the car reaches the top. Maximum episode length is 1000 steps. 
\end{itemize}

Furthermore, we provide the detailed dimensions of the states in our evaluated tasks in Table~\ref{tab:appendix-state-dimensions}.

\begin{table*}[h]
\centering
\small
\caption{The dimensions of the states in the evaluated tasks.}
\label{tab:appendix-state-dimensions}

\begin{tabular}{cc}
    \toprule
    Domain (Tasks) & Dimension \\
    \midrule
    Ant robot (\textit{AntStand}, \textit{AntFar}) & 105 \\
    Humanoid robot (\textit{HumanStand}, \textit{HumanKeep}) & 348 \\
    \textit{RobotReach} & 20 \\
    \textit{RobotPush}, \textit{RobotSlide} and \textit{RobotPickPlace} & 35 \\
    Atari games (\textit{MontezumaRevenge}, \textit{PitFall}, \textit{Frooger}, \textit{Solaris}, \textit{Freeway}) & $84\times84 = 7056$ \\
    \textit{MountainCar} & 2 \\
    \bottomrule
\end{tabular}
\end{table*}

\end{document}